\documentclass[conference]{IEEEtran}
\IEEEoverridecommandlockouts
\usepackage{cite}
\usepackage{amsmath,amssymb,amsfonts}
\usepackage{algorithmic}
\usepackage{graphicx}
\usepackage{textcomp}
\usepackage{xcolor}
\usepackage{url}
\def\BibTeX{{\rm B\kern-.05em{\sc i\kern-.025em b}\kern-.08em
    T\kern-.1667em\lower.7ex\hbox{E}\kern-.125emX}}

\usepackage{hyperref}
\usepackage{fancyhdr}
\usepackage{soul}
\usepackage{nicematrix}
\usepackage{colortbl}    
\usepackage{wrapfig,lipsum,booktabs}
\usepackage{mathtools}
\usepackage{multirow} % for multirow
\usepackage{makecell}
\usepackage{times}
\usepackage{epsfig}
\usepackage{dblfloatfix}

\usepackage{pifont}
\newcommand{\tabincell}[2]{\begin{tabular}{@{}#1@{}}#2\end{tabular}} % 
\usepackage{slashbox}
\usepackage{subfig}
\usepackage{enumitem}
\usepackage{tikz}

\definecolor{PurpleFG}{rgb}{0.54,0.35,0.71}
\definecolor{LightCyanBG}{rgb}{0.85,0.92,0.97}
\definecolor{CyanBG}{rgb}{0.71,0.85,0.93}
\definecolor{BlueBG}{rgb}{0,0.46,0.71}
\definecolor{LukeRed}{rgb}{0.91,0.38,0.33}
\definecolor{LLBlue}{rgb}{0.67,0.86,0.88}
\definecolor{LOrange}{rgb}{0.97,0.67,0.35}

\begin{document}
\IEEEoverridecommandlockouts
\IEEEpubid{\makebox[\columnwidth]{ 979-8-3503-7608-1/24\$31.00 \copyright2024 IEEE \hfill} \hspace{\columnsep}\makebox[\columnwidth]{ }}

\title{MG-Verilog: Multi-grained Dataset Towards Enhanced LLM-assisted Verilog Generation\vspace*{-0.3cm}}
% \author{Yongan Zhang, Zhongzhi Yu, Yonggan Fu, Cheng Wan,
% Yingyan (Celine) Lin}
% \email{celine.lin@gatech.edu}
% \affiliation{%
%   \institution{Georgia Institute of Technology}
%   % \streetaddress{P.O. Box 1212}
%   \city{Atlanta}
%   \state{Georgia}
%   % \postcode{43017-6221}
% }

\author{
\IEEEauthorblockN{Yongan Zhang, Zhongzhi Yu, Yonggan Fu, Cheng Wan, Yingyan (Celine) Lin}
\IEEEauthorblockA{
celine.lin@gatech.edu \\
Georgia Institute of Technology\\
Atlanta, Gerogia, USA
}\vspace*{-1cm}}

% \and
% \and
% \and
% \and
% \and

\maketitle
\begin{abstract}
Large Language Models (LLMs) have recently shown promise in streamlining hardware design processes by encapsulating vast amounts of domain-specific data. In addition, they allow users to interact with the design processes through natural language instructions, thus making hardware design more accessible to developers. However, effectively leveraging LLMs in hardware design necessitates providing domain-specific data during inference (e.g., through in-context learning), fine-tuning, or pre-training. Unfortunately, existing publicly available hardware datasets are often limited in size, complexity, or detail, which hinders the effectiveness of LLMs in hardware design tasks.
To address this issue, we first propose a set of criteria for creating high-quality hardware datasets that can effectively enhance LLM-assisted hardware design. Based on these criteria, we propose a Multi-Grained-Verilog (MG-Verilog) dataset, which encompasses descriptions at various levels of detail and corresponding code samples. To benefit the broader hardware design community, we have developed an open-source infrastructure that facilitates easy access, integration, and extension of the dataset to meet specific project needs.
Furthermore, to fully exploit the potential of the MG-Verilog dataset, which varies in complexity and detail, we introduce a balanced fine-tuning scheme. This scheme serves as a unique use case to leverage the diverse levels of detail provided by the dataset. Extensive experiments demonstrate that the proposed dataset and fine-tuning scheme consistently improve the performance of LLMs in hardware design tasks.

\end{abstract}
\section{Introduction}
\label{sec:intro}

Large Language Models (LLMs) have recently emerged as a promising approach to streamline hardware design processes~\cite{thakur2023verigen,liu2023verilogeval,fu2023gpt4aigchip,yan2023viability,he2023chateda,blocklove2023chip}. By encapsulating vast amounts of domain-specific data and enabling users to interact with the design processes through natural language prompts, LLMs have the potential to make hardware design more accessible to a broader range of developers. This increased accessibility can foster innovation and accelerate the development of new hardware solutions, as it allows developers with varying levels of expertise to contribute to design processes.

% Despite the great potential of LLMs, existing state-of-the-art (SOTA) general LLMs, such as OpenAI's GPT-4~\cite{openai2023gpt4}, may be limited in their ability to generate practical hardware designs. For example, they might generate non-synthesizable or non-functional hardware source code~\cite{fu2023gpt4aigchip}. To address this limitation, recent studies suggest that incorporating additional domain-specific data is crucial for enhancing LLMs' performance in hardware design tasks, using techniques across the scopes of LLM inference, fine-tuning, or pre-training. Specifically, one approach to improve LLMs' hardware design capabilities is to provide them with additional relevant design examples during inference-only generation. GPT4AIGChip~\cite{fu2023gpt4aigchip} demonstrated that this method can significantly enhance the quality of generated High-Level Synthesis (HLS) hardware code. Another approach is to fine-tune LLMs on carefully curated hardware design datasets, such as VerilogEval~\cite{liu2023verilogeval}, which has been shown to improve LLMs' performance in generating Verilog code. Alternatively, LLMs can also be pre-trained on diverse datasets from various hardware design domains to specialize in general hardware design concepts, as exemplified by ChipNemo~\cite{liu2023chipnemo}, leading to improved general performance across a range of hardware design tasks.
Despite the great potential of LLMs, existing state-of-the-art (SOTA) general LLMs, e.g., OpenAI's GPT-4~\cite{openai2023gpt4}, are still limited in their ability to generate practical hardware designs. For example, they might generate non-synthesizable or non-functional hardware source code~\cite{fu2023gpt4aigchip}. To address this limitation, recent studies suggest that incorporating additional domain-specific data is crucial for enhancing LLMs' performance in hardware design tasks, using techniques across the scopes of LLM inference, fine-tuning, or pre-training. Specifically, one approach to improve LLMs' hardware design capabilities is to provide them with additional relevant design examples during inference-only generation, e.g., GPT4AIGChip~\cite{fu2023gpt4aigchip}. It has been shown that this method can significantly enhance the quality of generated High-Level Synthesis (HLS) hardware code. Another approach is to fine-tune LLMs on carefully curated hardware design datasets, e.g., VerilogEval~\cite{liu2023verilogeval}, which has been shown to improve LLMs' performance in generating Verilog code. Alternatively, LLMs can also be pre-trained on diverse datasets from various hardware design domains to specialize in general hardware design concepts, as exemplified by ChipNemo~\cite{liu2023chipnemo}, leading to improved general performance across a range of hardware design tasks.

Although the aforementioned approaches show promise in enhancing LLMs' performance in hardware design tasks, their progress can be hindered by the limitations of current publicly available hardware design datasets. As we will later analyze, the size, complexity, and detail granularity of datasets are essential factors for improving LLMs' performance. However, existing datasets often fall short in one or more of these aspects. Some datasets, e.g., those used in~\cite{lu2023rtllm,thakur2022benchmarking,blocklove2023chip}, contain only a small number of data points (e.g., under 2e2), which are only suitable for benchmarking the LLMs' task performance but is insufficient for effectively fine-tuning LLMs. Other datasets, like those employed in~\cite{liu2023verilogeval,thakur2023verigen}, can be simplistic, either lacking important features (e.g., code samples containing multiple module instantiations and aligned descriptions) or providing only high-level descriptions for each code piece. This simplicity can limit the fine-tuned LLMs' generalization performance when faced with diverse user instructions, thus reducing their effectiveness.

To address the limitations of existing datasets and unlock the full potential of LLM fine-tuning and in-context learning for hardware design tasks, we propose a Multi-Grained-Verilog (MG-Verilog) dataset. This dataset includes hardware descriptions at different levels of detail and their corresponding Verilog code samples with varying design complexity. These features make it suitable for both inference and fine-tuning stages of LLMs to enhance their performance in hardware design tasks. Our main contributions can be summarized as follows:

\begin{itemize}
    % \item We introduce a set of essential criteria for high-quality hardware datasets that can be effectively utilized by LLM-assisted hardware design techniques. These criteria serve as a guide for the development of future datasets in this domain.
    \item We introduce a set of essential criteria for high-quality hardware datasets that can be effectively utilized by LLM-assisted hardware design techniques. These criteria can serve as a guide for the development of future datasets in this domain.

    % \item We present the open-source MG-Verilog dataset\footnote{\url{https://anonymous.4open.science/r/mg-verilog-B254/}}, which meets the aforementioned criteria. Additionally, we provide the necessary infrastructure for users to access, integrate, and extend the dataset for their specific project needs, promoting collaboration and facilitating further research in this area.
    \item We present an open-source MG-Verilog dataset\footnote{\url{https://github.com/luke-avionics/mg-verilog}}, which meets the aforementioned criteria. Additionally, we provide the necessary infrastructure for users to access, integrate, and extend the dataset for their specific project needs, promoting collaboration and facilitating further research in this area.
    
    % \item We demonstrate a unique use case of the MG-Verilog dataset by proposing a balanced fine-tuning scheme that leverages the diverse levels of detail provided by the dataset. This scheme showcases the potential of the dataset to enable novel approaches in LLM-assisted hardware design.
    \item We demonstrate a unique use case of the MG-Verilog dataset by proposing a balanced fine-tuning scheme that leverages the diverse levels of detail provided by the dataset. This scheme validates and showcases the potential of the dataset to enable novel approaches in LLM-assisted hardware design.
    
    \item Extensive experiments show that LLMs fine-tuned with our MG-Verilog dataset outperform those trained on datasets from other sources in terms of both code implementation accuracy and the sophistication of generated hardware designs. These results highlight the effectiveness of our dataset in enhancing LLMs' performance for hardware design tasks.
\end{itemize}
\section{Criteria for Datasets in LLM-assisted Hardware Design} 
\label{sec:background}
To create a high-quality dataset for LLM-assisted hardware design, we first establish design criteria to guide the development of the MG-Verilog dataset. 

% \textbf{Sufficient dataset size.} It is crucial for both training (i.e., domain-specific pre-training or fine-tuning) and inference (i.e., in-context learning) of LLMs. A larger dataset provides diverse examples for improved generalization performance during training~\cite{liu2023verilogeval,liu2023chipnemo} and enables effective techniques such as Retrieval-Augmented-Generation (RAG) for enhanced generation quality during inference~\cite{fu2023gpt4aigchip}.
\textbf{Sufficient dataset size.} This is crucial for both training (i.e., domain-specific pre-training or fine-tuning) and inference (i.e., in-context learning) of LLMs. A larger dataset provides diverse examples for improved generalization performance during training~\cite{liu2023verilogeval,liu2023chipnemo} and enables effective techniques such as Retrieval-Augmented-Generation (RAG) for enhanced generation quality during inference~\cite{fu2023gpt4aigchip}.

\textbf{Accurate code-description pairs.} Each code sample needs to be correct, functional, and associated with a precise description of its functionality. Inaccuracies or ambiguity can mislead LLMs during fine-tuning or pre-training and lead to erroneous code generation during inference.

\textbf{Varied description detail levels.} They are necessary to address two challenges. Datasets with only high-level descriptions may not provide sufficient detail for accurate code generation or effective LLM training (i.e., fine-tuning or pre-training), especially for complex designs. Conversely, datasets dominated by detailed descriptions may limit practical utility, as LLMs trained on such datasets might require users to provide elaborated prompts, which can be as labor-intensive as coding from scratch.
Hence, an effective dataset should incorporate both high-level and detailed descriptions in a proper balance. In particular, high-level descriptions can facilitate user-friendly LLM interactions, while detailed descriptions are crucial for enabling LLMs to create complex designs, offering in-depth guidance for LLMs during training, or serving as a comprehensive reference during inference.

\textbf{Extensibility and integrability for future development.} A high-quality hardware dataset should be designed with the research community in mind, allowing for easy extension and integration into various projects. The rapidly evolving nature of hardware design necessitates a dataset that can adapt to the latest trends and requirements.
Moreover, the vast scope of hardware design means that different developers may have specific focused areas, making it challenging for a single organization to cover all possible scenarios in a one-time effort. To address this issue, the dataset should be structured in a way that encourages researchers to contribute to its growth and adapt it to their specific needs, fostering collaboration within the research community and ensuring its relevance and utility. This approach not only benefits individual projects but also contributes to the overall advancement of LLM-assisted hardware design methodologies.

% However, existing hardware datasets often fall short in one or more of these aspects, as we will later analyze in Sec.~\ref{sec:related_works}. To address these limitations and unlock the full potential of LLM fine-tuning and in-context learning, we propose the Multi-Grained-Verilog (MG-Verilog) dataset, which meets the aforementioned criteria.

\begin{figure}
    % \subfloat[MG-Verilog dataset structure]{
    %     \includegraphics[width=1\linewidth]{Figs/pot-structure.pdf}
    % }\vspace{-1em}
    \includegraphics[width=1\linewidth]{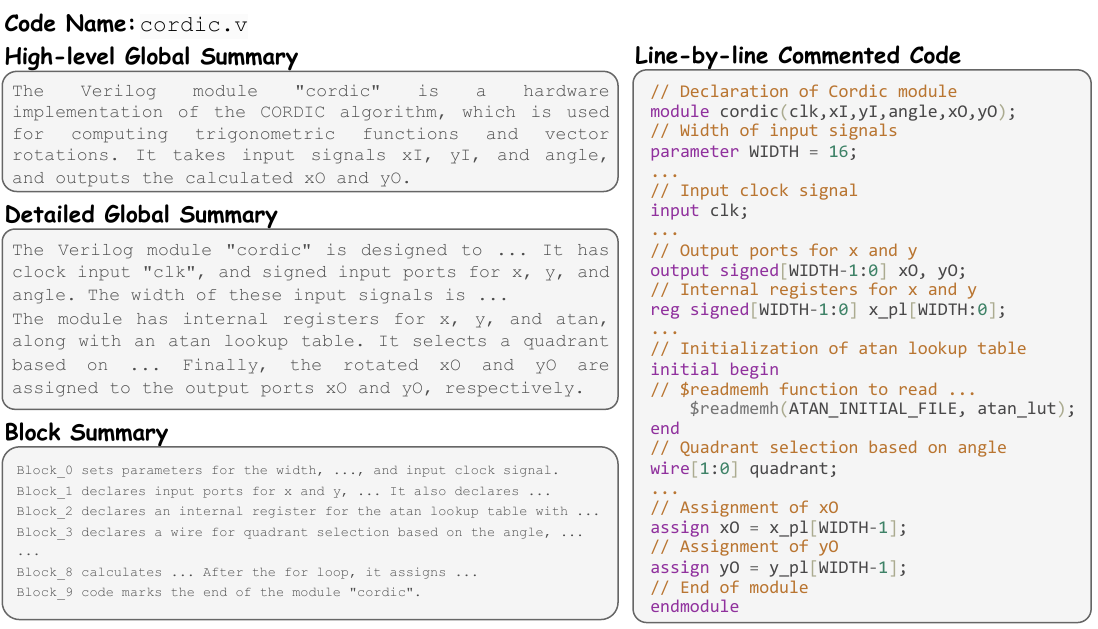}
    \vspace{-1.5em}
    \caption{Illustrating the proposed MG-Verilog dataset structure and examples of varying levels of detail.}
    \label{fig:pot}
    \vspace{-1em}
\end{figure}

\section{The Proposed MG-Verilog Dataset} \label{sec:method_dataset}

\subsection{Dataset Overview} \label{sec:dataset_overview}
The MG-Verilog dataset consists of over 11,000 Verilog code samples and their corresponding natural language descriptions, serving as the desired outputs and test inputs for various LLM-assisted hardware design tasks, such as Verilog code generation. 
% Sec.~\ref{sec:training_sample_ablation} shows that the total dataset size is sufficient enough for fine-tuning LLMs for the basic Verilog code generation tasks~\cite{liu2023verilogeval}.
% The dataset is constructed using open-source Verilog code from GitHub~\cite{github} repositories, leveraging BigQuery's snapshots~\cite{BigQuery,thakur2023verigen}. Detailed statistics of the dataset are provided in Sec.~\ref{sec:dataset_statistics}.

\subsection{Dataset Construction} 
\label{sec:dataset_construction}
The construction of the MG-Verilog dataset involves several steps to ensure the quality and usability of the data.

\subsubsection{Data Collection and Preprocessing} \label{sec:dataset_preprocessing}
Raw source code from open-source repositories is collected and preprocessed to ensure correctness. Adapting from VerilogEval~\cite{liu2023verilogeval}, we use Pyverilog~\cite{Takamaeda:2015:ARC:Pyverilog} to parse the raw Verilog code and exclude code samples containing syntax errors. Deduplication techniques are applied to remove redundant code samples. Additionally, dependencies of the code samples are extracted, i.e., sub-modules of multi-module code samples are identified and recorded as metadata to facilitate research on techniques such as few-shot learning and RAG for generating multi-module Verilog code.

\subsubsection{Description Generation} 
\label{sec:description_generation}
Natural language descriptions are appended to the code samples using an approach similar to VerilogEval~\cite{liu2023verilogeval}, leveraging LLMs' superior natural language generation capabilities. In addition to simple high-level descriptions for each code piece, varying levels of detailed descriptions aligned with the code complexity are provided, as detailed in Sec.~\ref{sec:dataset_structure}.

\subsection{Multi-grained Dataset Structure}
\label{sec:dataset_structure}
% To strike a balance between design generation accuracy and user-friendliness, we introduce the multi-grained data structure, which encompasses descriptions at various levels of detail in order to satisfy the third criterion in Sec.~\ref{sec:background}. As depicted in Fig.~\ref{fig:pot}, this structure organizes hardware code descriptions, ranging from high-level summaries to detailed, line-by-line comments.
% The multi-grained structure is designed to mimic the learning and design processes of human designers. The objective is to simplify the learning curve for using the dataset and, as demonstrated later, to better leverage the strengths of LLMs for enhanced description generation accuracy. 
% Specifically, the multi-grained structure mirrors the typical two phases experienced by human designers. In the learning phase, a hardware designer starts with the basic syntax and semantics of the design language, gradually advancing to apply this knowledge to design higher-level hardware modules. Conversely, in the design phase, the process begins with high-level architectural planning for the entire design, followed by a detailed, step-by-step implementation.
To strike a balance between design generation accuracy and user-friendliness, we adopt a multi-grained data structure, which encompasses descriptions at various levels of detail in order to satisfy the third criterion in Sec.~\ref{sec:background}. As depicted in Fig.~\ref{fig:pot}, this structure organizes hardware code descriptions, ranging from high-level summaries to detailed, line-by-line comments.
The multi-grained structure is designed to mimic the learning and design processes of human designers. The objective is to simplify the learning curve for using the dataset and, as demonstrated later, to better leverage the strengths of LLMs for enhanced description generation accuracy. 
Specifically, the multi-grained structure mirrors the typical two phases experienced by human designers. In the learning phase, a hardware designer starts with the basic syntax and semantics of the design language, gradually advancing to apply this knowledge to design higher-level hardware modules. Conversely, in the design phase, the process begins with high-level architectural planning for the entire design, followed by a detailed, step-by-step implementation.

\begin{figure}
    \vspace{-0.7em}
    \includegraphics[width=1\linewidth]{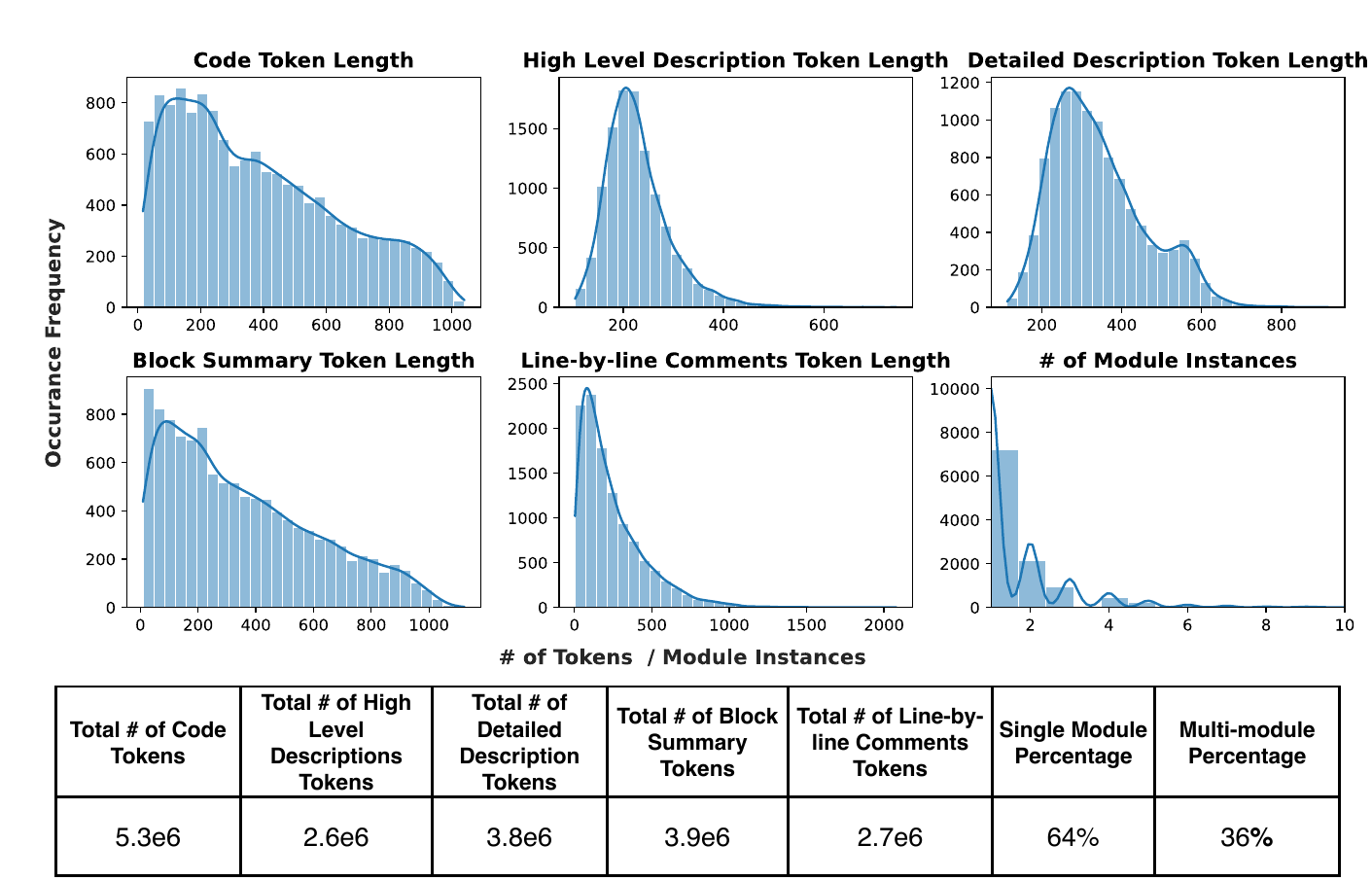}
    \vspace{-1.5em}
    \caption{The detailed statistics of the MG-Verilog dataset, using the tokenizer from the GPT-3.5-Turbo model~\cite{gpt35}.}
    \label{fig:dataset_stats}
    \vspace{-1.5em}
\end{figure}

\subsection{Detailed Statistics of the Dataset} \label{sec:dataset_statistics}
Fig.~\ref{fig:dataset_stats} presents detailed statistics of the MG-Verilog dataset, illustrating the distribution of token length for both the code and varying levels of descriptions. The complexity of the code samples is also reflected in the distribution of the number of module instances. The dataset shows a wide range of natural language description details and code complexities, making it suitable for diverse LLM-assisted hardware design tasks.

\subsection{Dataset Access and Extension Instructions} \label{sec:dataset_access}
The MG-Verilog dataset is publicly available and packaged in the standard HuggingFace Datasets format~\cite{huggingface_datasets} for easy access and integration. Each dataset entry contains the following fields: \textit{code}, \textit{high-level summaries}, \textit{detailed summaries}, \textit{block-level summaries}, \textit{line-by-line comments}, and \textit{metadata}. The metadata field currently includes the module dependencies of the code samples.
The MG-Verilog dataset is open-sourced from raw data collection to the final dataset construction in a modular manner for straightforward extension.
The demonstrated balanced fine-tuning use case is also provided as a reference. 
% The entire framework is designed as a modular system and is completely based on open-source tools, allowing users to easily extend the dataset or integrate it into their specific projects using the entire pipeline or individual components.

\section{Dataset Unique Use Case: A Balanced Fine-tuning Scheme}

\label{sec:method_finetune}
In this section, we show a unique use case of our proposed MG-Verilog dataset. Specifically, we introduce a balanced fine-tuning scheme to fully harness the diverse levels of detail provided by our MG-Verilog dataset.

\textbf{The challenge to address.} The ultimate goal of fine-tuning is to generate hardware code solely from high-level design descriptions. However, challenges arise when determining the type of descriptions to be used for fine-tuning. On the one hand, fine-tuning with only simple high-level descriptions may not provide LLMs with sufficient information to generate code for complex designs. On the other hand, exclusively relying on detailed descriptions could hinder LLMs' ability to respond to more high-level user instructions.

\textbf{Our balanced fine-tuning scheme.}
To tackle the aforementioned challenge, we present a balanced fine-tuning scheme that randomly selects training samples with varying levels of descriptions from the MG-Verilog dataset in each fine-tuning iteration. The aim is to achieve a balance when imparting knowledge of both global and local code semantics to LLMs.

% Specifically, this approach mixes training samples with three different levels outlined in Fig.~\ref{fig:pot}, including:
% \underline{(1)} high-level global summaries, which are simple overarching summaries of the entire code, instructing LLMs in understanding and generating code from brief instructions;
% \underline{(2)} detailed global summaries, which offer in-depth information to guide LLMs in complex code generation; and
% \underline{(3)} block-level summaries, which provide detailed information for individual code blocks.
% Each type of training sample is formatted into a custom prompt template and paired with the corresponding code samples. It is worth noting that line-by-line comments are excluded from the training samples as they overly focus on local semantics. The effectiveness of this balanced fine-tuning scheme is demonstrated in Sec.~\ref{sec:different_eval_settings}.
\section{Experimental Results}
\label{sec:experiments}
\subsection{Experiment Setup}
\label{sec:setup}

\indent\textbf{Dataset generation.} 
The primary model for generating descriptions is LLaMA2-70B-Chat. GPT-3.5-turbo serves as an automated backup for scenarios where the maximum token limit is exceeded.
Based on empirical testing, we set the temperature to 0.7 and $top\_p$ to 0.95, maintaining other hyperparameters at their default values for the best quality.

\begin{table}[t]\centering
    \vspace{-0em}
    \caption{Comparison across fine-tuning and evaluation data formats using the CodeLLaMA-7B-Instruct model. The table columns indicate the data formats used for fine-tuning, while the rows show the formats used during evaluation. Performance is color-coded for clarity: warm colors (red and orange) indicate high performance, while cool colors (light blue and blue) denote lower performance. The color gradient from best to worst performance is as follows: \textbf{red (highest)}, \textbf{orange}, \textbf{light blue}, and \textbf{blue (lowest)}. A notation of \textit{H}, \textit{MH}, \textit{ML}, and \textit{L} is used to indicate high, medium (high/low), and low performance, respectively, for better visual clarity.}
    \vspace{-0em}
    \label{tab:train_inference_settings}
    \scriptsize
    \resizebox{\linewidth}{!}{
        \begin{tabular}{l||c|c|c|c}
        \toprule[2pt]
        \multicolumn{5}{c}{\textbf{Pass@1}} \\
        \midrule
        % Techniques &Workload  & \tabincell{c}{Throughput\\(FPS)} & \tabincell{c}{Norm.\\Energy Eff.} \\\hline
        \backslashbox{Evaluate}{Fine-tune} & \tabincell{c}{\textcolor{blue}{\textbf{MG-Verilog}}\\\textcolor{blue}{\textbf{Balanced Fine-tune}}} &  \tabincell{c}{High Level \\ Global Summaries} &  \tabincell{c}{Detailed \\Global Summaries} &  Block Summaries\\
        \midrule
        \tabincell{c}{High Level\\Global Summaries}   & \cellcolor{LukeRed!45}45.2 \; \textit{H} & \cellcolor{LLBlue!65}42.4 \; \textit{ML}&\cellcolor{LOrange!65}44.8 \; \textit{MH}& \cellcolor{BlueBG!35}40.3 \; \textit{L}\\
        \midrule
        \tabincell{c}{Detailed\\Global Summaries}   & \cellcolor{LOrange!65}52.7  \; \textit{MH}& \cellcolor{LLBlue!65}50.8 \; \textit{ML}&\cellcolor{LukeRed!45}54.5 \; \textit{H}& \cellcolor{BlueBG!35}46.3 \; \textit{L}\\
        \midrule
        \tabincell{c}{Block Summaries}   & \cellcolor{LOrange!65}51.1  \; \textit{MH}& \cellcolor{LLBlue!65}41.8 \; \textit{ML}&\cellcolor{BlueBG!35}40.0 \; \textit{L}& \cellcolor{LukeRed!45}52 \; \textit{H}\\
        \midrule
        \multicolumn{5}{c}{\textbf{Pass@5}} \\
        \midrule
        \backslashbox{Evaluate}{Fine-tune}  & \tabincell{c}{\textcolor{blue}{\textbf{MG-Verilog}}\\\textcolor{blue}{\textbf{Balanced Fine-tune}}} & \tabincell{c}{High Level \\ Global Summaries} &   \tabincell{c}{Detailed \\Global Summaries}  &  Block Summaries\\
        \midrule
        \tabincell{c}{High Level\\Global Summaries}   & \cellcolor{LukeRed!45}52.2 \; \textit{H}& \cellcolor{LLBlue!65}48.1 \; \textit{ML}&\cellcolor{LOrange!65}49.9  \; \textit{MH}& \cellcolor{BlueBG!35}44.0 \; \textit{L}\\
        \midrule
        \tabincell{c}{Detailed\\Global Summaries}   & \cellcolor{LOrange!65}58.5  \; \textit{MH}& \cellcolor{LOrange!65}58.5  \; \textit{MH}&\cellcolor{LukeRed!45}59.7 \; \textit{H}& \cellcolor{BlueBG!35}52.2 \; \textit{L}\\
        \midrule
        \tabincell{c}{Block Summaries}   & \cellcolor{LOrange!65}56.2 \; \textit{MH}&  \cellcolor{LLBlue!65}52.5 \; \textit{ML}&\cellcolor{BlueBG!35}46.3 \; \textit{L}& \cellcolor{LukeRed!45}60 \; \textit{H}\\
        \midrule
        \multicolumn{5}{c}{\textbf{Pass@10}} \\
        \midrule
        \backslashbox{Evaluate}{Fine-tune} & \tabincell{c}{\textcolor{blue}{\textbf{MG-Verilog}}\\\textcolor{blue}{\textbf{Balanced Fine-tune}}} &  \tabincell{c}{High Level \\ Global Summaries} &   \tabincell{c}{Detailed \\Global Summaries}  &  Block Summaries\\
        \midrule
        \tabincell{c}{High Level\\Global Summaries}   & \cellcolor{LukeRed!45}55.2 \; \textit{H}&  \cellcolor{LLBlue!65}49.7 \; \textit{ML}&\cellcolor{LOrange!65}51.8 \; \textit{MH}& \cellcolor{BlueBG!35}45.6 \; \textit{L}\\
        \midrule
        \tabincell{c}{Detailed\\Global Summaries}   & \cellcolor{LOrange!65}60.9 \; \textit{MH}& \cellcolor{LukeRed!45}61.5 \; \textit{H}& \cellcolor{LLBlue!65}60.1 \; \textit{ML}& \cellcolor{BlueBG!35}53.1 \; \textit{L}\\
        \midrule
        \tabincell{c}{Block Summaries}   & \cellcolor{LOrange!65}58.0 \; \textit{MH}&  \cellcolor{LLBlue!65}54.5 \; \textit{ML}&\cellcolor{BlueBG!35}47.6 \; \textit{L}& \cellcolor{LukeRed!45}63 \; \textit{H}\\
        \midrule
        \bottomrule[2pt]
        \end{tabular}
    }
    \vspace{-2em}
    \end{table}

\textbf{Fine-tuning and inference.} 
CodeLLaMA-7B-Instruct is chosen as the primary model for hardware code generation due to its superior coding performance and small model size. For fine-tuning it on our dataset, the fine-tuning approach is based on QLoRA~\cite{dettmers2023qlora}, using its default training settings to demonstrate our delivered dataset's effectiveness.
The fine-tuned model is evaluated using 143 Verilog coding questions from the benchmark in~\cite{liu2023verilogeval}, excluded from the training set.

\textbf{Hardware evaluation and metrics.}  The validity of each generated design is tested by compiling it and checking against its RTL simulation results in pre-defined testbench cases. We employ unbiased pass@1, pass@5, and pass@10 metrics, calculated from 20 generation runs, as established in~\cite{liu2023verilogeval}.

\vspace{-0.5em}
\subsection{Ablation Study on Different Evaluation Settings}
\label{sec:different_eval_settings}
In this section, we explore the performance of fine-tuned models using varying data formats in both the training and evaluation phases. 
Although high-level global summaries are the most user-friendly data format, their ambiguity often results in a lack of detailed information necessary for precise code generation. In some cases, detailed global summaries can actually be more advantageous for expert users who have a deep understanding of code structures. Consequently, an ideal RTL code generation dataset would facilitate consistent model performance across a range of input instruction complexities. 

\textbf{Observations and analysis.}
Tab.~\ref{tab:train_inference_settings} provides insights into these findings. 
Notably, we can observe: \underline{(1)} Models fine-tuned with the MG-Verilog dataset exhibit the most robust performance in all tested evaluation settings. Specifically, while different evaluation settings tend to bias the fine-tuning setting that aligns with them, models fine-tuned with the MG-Verilog dataset consistently rank in the top two positions when compared to other baselines. In contrast, other baselines may perform well only under their aligned evaluation settings and notably under-perform in other evaluation settings. \underline{(2)} Training exclusively with either overly detailed or overly high-level data can result in decreased performance, indicating the importance of having balanced training data. 
Specifically, Tab.~\ref{tab:train_inference_settings} reveals that, apart from the MG-Verilog dataset, models trained with detailed global summaries yield the highest pass rates. These summaries strike a balance between the generality of high-level global summaries and the specificity of block summaries.

\begin{figure}
    \centering
    \includegraphics[width=1\linewidth]{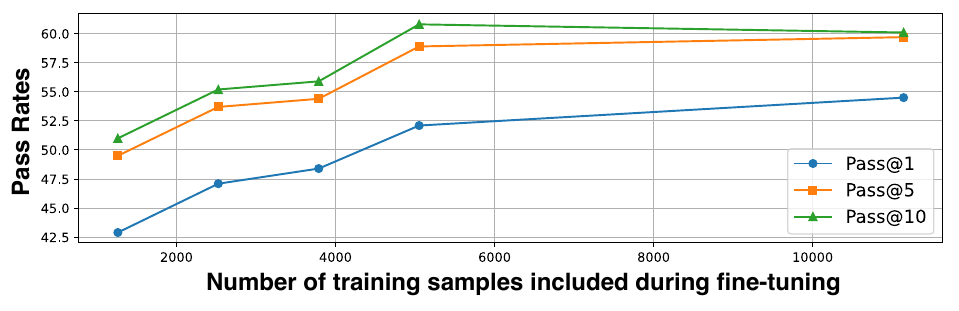}
        \vspace{-2em}
        \caption{Pass rates of the generated RTL code from fine-tuned CodeLLaMA-7B-Instruct model using different numbers of training samples. Here only detailed global summaries of the code are used during the fine-tuning.}
    \label{fig:train_samples}
    \vspace{-1.5em}
\end{figure}

\vspace{-0.5em}
\subsection{Ablations on the Number of Training Samples} 
\label{sec:training_sample_ablation}
We further examine how the quantity of training samples affects the performance of models fine-tuned for RTL code generation tasks. As illustrated in Fig.~\ref{fig:train_samples}, there is a clear trend where the model's performance improves with an increase in the number of training samples. However, we also note a diminishing returns phenomenon. Specifically, the performance gains from additional training samples decrease as the total number of samples grows. This trend could be attributed to either the limited diversity in the raw source code or the potential need for more optimal hyperparameter tuning and model configurations. These aspects, being orthogonal to the dataset structure proposed, are left for future exploration.

% \subsection{Cost of Dataset Generation and Fine-tuning}
% Prove it that it's viable even for the academia to generate such dataset and fine-tune the model.

% \subsection{Human Efforts Saving on Dataset Preparation and Effectiveness of Guiding developers}

\section{Related Work}
\label{sec:related_works}
% LLMs have been applied in various stages of the hardware design process, including verification~\cite{srikumarfast}, security flaw detection~\cite{paria2023divas}, and code generation~\cite{thakur2023verigen,liu2023verilogeval,fu2023gpt4aigchip,chang2023chipgpt}. However, their performance is still limited due to insufficient exposure to hardware data during pretraining~\cite{fu2023gpt4aigchip,chang2023chipgpt}. Some studies~\cite{thakur2023verigen,liu2023verilogeval,lu2023rtllm} have tried to rectify this by supplying more hardware code samples and fine-tuning the LLMs. Yet, the datasets used are either too small~\cite{lu2023rtllm} or overly simplistic~\cite{liu2023verilogeval,thakur2023verigen}, which hinder effective fine-tuning of LLMs. The MG-Verilog dataset addresses this issue by providing an open-sourced, high-quality dataset, essential for optimizing LLM fine-tuning and in-context learning.
LLMs have been applied in various stages of the hardware design process, including verification~\cite{srikumarfast}, security flaw detection~\cite{paria2023divas}, and code generation~\cite{thakur2023verigen,liu2023verilogeval,fu2023gpt4aigchip,chang2023chipgpt}. However, their performance is still limited due to insufficient exposure to hardware data during pretraining~\cite{fu2023gpt4aigchip,chang2023chipgpt}. Some studies~\cite{thakur2023verigen,liu2023verilogeval,lu2023rtllm} have tried to rectify this by supplying more hardware code samples and fine-tuning the LLMs. Yet, the datasets used are still either too small~\cite{lu2023rtllm} or overly simplistic~\cite{liu2023verilogeval,thakur2023verigen}, which hinder effective fine-tuning of LLMs. Our MG-Verilog dataset addresses this issue by providing an open-sourced, high-quality dataset, essential for optimizing LLM fine-tuning and in-context learning.

\section{Conclusion}
% In this work, we have addressed the limitations of existing datasets for LLM-assisted hardware design by proposing the Multi-Grained-Verilog (MG-Verilog) datasets. The MG-Verilog dataset includes hardware descriptions at different levels of detail and their corresponding Verilog code samples for more generic use cases.
% We have demonstrated the effectiveness of the dataset through a balanced fine-tuning scheme. Extensive experiments show that LLMs fine-tuned with the MG-Verilog dataset outperform those trained on other datasets in terms of Verilog code generation accuracy. By making the dataset openly accessible and providing the necessary infrastructure for integration and extension, we promote collaboration within the research community and facilitate future advancements in LLM-assisted hardware design.
In this work, we aim to mitigate the limitations of existing datasets for LLM-assisted hardware design by proposing the open-sourced Multi-Grained-Verilog (MG-Verilog) dataset. The MG-Verilog dataset features hardware descriptions at different levels of detail and their corresponding Verilog code samples for more generic use cases.
We have demonstrated the effectiveness of the dataset through a balanced fine-tuning scheme. Extensive experiments show that LLMs fine-tuned with the MG-Verilog dataset outperform those trained on other datasets in terms of Verilog code generation accuracy. 
% By making the dataset openly accessible and providing the necessary infrastructure for integration and extension, we promote collaboration within the research community and facilitate future advancements in LLM-assisted hardware design.

% \vspace{-0.5em}
\section{Acknowledgments}
The work is supported by the National Science Foundation (NSF) through the RTML funding (Award number: 2400511)
, an NSF CAREER award (Award number: 2345577), and CoCoSys, one of the seven centers in JUMP 2.0, a Semiconductor Research Corporation (SRC) program sponsored by DARPA.

\bibliographystyle{IEEEtranS}
\bibliography{ref}
\end{document}